\documentclass[sigconf]{aamas}
\usepackage{balance} 
\usepackage{amsmath}

\usepackage{graphicx}
\usepackage{multirow}
\usepackage{url}
\usepackage[ruled,vlined,linesnumbered]{algorithm2e}
\usepackage{verbatim}

\DeclareMathOperator{\argmin}{arg\,min}

\title[Robust Multi-Agent Pickup and Delivery with Delays]{Robust Multi-Agent Pickup and Delivery with Delays}
\setcopyright{none}
\renewcommand\footnotetextcopyrightpermission[1]{} 
\author{Giacomo Lodigiani}
\affiliation{
  \institution{Politecnico di Milano}
  \city{Milan}
  \country{Italy}}
\email{giacomo.lodigiani@mail.polimi.it}

\author{Nicola Basilico}
\affiliation{
  \institution{Università degli Studi di Milano}
  \city{Milan}
  \country{Italy}}
\email{nicola.basilico@unimi.it}

\author{Francesco Amigoni}
\affiliation{
  \institution{Politecnico di Milano}
  \city{Milan}
  \country{Italy}}
\email{francesco.amigoni@polimi.it}

\begin{abstract}
Multi--Agent Pickup and Delivery (MAPD) is the problem of computing collision-free paths for a group of agents such that they can safely reach delivery locations from pickup ones. These locations are provided at runtime, making MAPD a combination between classical Multi--Agent Path Finding (MAPF) and online task assignment. Current algorithms for MAPD do not consider many of the practical issues encountered in real applications: real agents often do not follow the planned paths perfectly, and may be subject to delays and failures. In this paper, we  study the problem of MAPD with \textit{delays}, and we present two solution approaches that provide robustness guarantees by planning paths that limit the effects of imperfect execution. In particular, we introduce two algorithms, $k$--TP and $p$--TP, both based on a decentralized algorithm typically used to solve MAPD, Token Passing (TP), which offer deterministic and probabilistic guarantees, respectively. Experimentally, we compare our algorithms against a version of TP enriched with online replanning. $k$--TP and $p$--TP provide robust solutions, significantly reducing the number of replans caused by delays, with little or no increase in solution cost and running time.
\end{abstract}

\newcommand{\BibTeX}{\rm B\kern-.05em{\sc i\kern-.025em b}\kern-.08em\TeX}

\begin{document}
\maketitle 

\section{Introduction}
In Multi--Agent Pickup and Delivery (MAPD)~\cite{Ma}, a set of agents must jointly plan collision--free paths to serve pickup--delivery tasks that are submitted at runtime. MAPD combines a task-assignment problem, where agents must be assigned to pickup--delivery pairs of locations, with Multi--Agent Path Finding (MAPF)~\cite{Stern}, where collision--free paths for completing the assigned tasks must be computed. A particularly challenging feature of MAPD problems is that they are meant to be cast into dynamic environments for long operational times. In such settings, tasks can be submitted at any time in an online fashion.

Despite studied only recently, MAPD has a great relevance for a number of real--world application domains. Automated warehouses, where robots continuously fulfill new orders, arguably represent the most significant industrial deployments~\cite{Wurman}. Beyond logistics, MAPD applications include also the coordination of teams of service robots~\cite{Veloso} or fleets of autonomous cars, and the automated control of non--player characters in video games~\cite{Ma_games}.

Recently, the MAPF community has focused on resolution approaches that can deal with real--world--induced relaxations of some idealistic assumptions usually made when defining the problem. A typical example is represented by the assumption that planned paths are executed without errors. In reality, execution of paths might be affected by delays and other issues that can hinder some of their expected properties (e.g., the absence of collisions). One approach is to add online adaptation to offline planning, in order to cope with situations where the path execution incurs in errors~\cite{ma2017multi}. Despite being reasonable, this approach is not always desirable in real robotic applications. Indeed, replanning can be costly in those situations where additional activities in the environment are conditioned to the plans the agents initially committed to. In other situations, replanning cannot even be possible: think, as an example, to a centralized setting where robots are no more connected to the base station when they follow their computed paths. This background motivated the study of \emph{robustness}~\cite{Atzmon,Atzmon_k_JAIR,Ma}, generally understood as the capacity, guaranteed at planning time, of agents' paths to withstand unexpected runtime events. In our work, we focus on robustness in the long--term setting of MAPD, where it has not been yet consistently studied.

Specifically, in this paper, we study the robustness of MAPD to the occurrence of \emph{delays}.
To do so, we introduce a variant of the problem that we call \textit{MAPD with delays} (\textit{MAPD--d} for short). In this variant, like in standard MAPD, agents must be assigned to tasks (pickup--delivery locations pairs), which may continuously appear at any time step, and collision--free paths to accomplish those tasks must be planned. However, during path execution, delays can occur at arbitrary times causing one or more agents to halt at some time steps, thus slowing down the execution of their planned paths. We devise a set of algorithms to compute robust solutions for MAPD--d. The first one is a baseline built from a decentralized MAPD algorithm, Token Passing (TP), to which we added a mechanism that replans in case collisions caused by delays are detected when following planned paths. TP is able to solve well--formed MAPD problem instances~\cite{Koenig}, and we show that, under some assumptions, the introduction of delays in MAPD--d does not affect well--formedness. We then propose two new algorithms, $k$--TP and $p$--TP, which adopt the approach of robust planning, computing paths that limit the risk of collisions caused by potential delays. $k$--TP returns solutions with deterministic guarantees about robustness in face of delays ($k$--robustness), while solutions returned by $p$--TP have probabilistic robustness guarantees ($p$--robustness). We compare the proposed algorithms by running experiments in simulated environments and we evaluate the trade--offs offered by different levels and types of robustness.

In summary, the main contributions of this paper are: the introduction of the MAPD--d problem and the study of some of its properties (Section~\ref{mapd_delays}), the definition of two algorithms ($k$--TP and $p$--TP) for solving MAPD--d problems with robustness guarantees (Section~\ref{S:algorithms}), and their experimental evaluation that provides insights about how robustness and solution cost can be balanced (Section~\ref{S:experiments}).

\section{Preliminaries and Related Work}\label{Related Works}
In this section, we discuss the relevant literature related to our work and we introduce the formal concepts we will build upon in the following sections.

A basic MAPF problem assigns a start--goal pair of vertices on a graph $G = (V, E)$ to each agent from a set $A = \{ a_{1}, a_{2}, \ldots, a_{\ell} \}$ and is solved by a minimum--cost discrete--time set of paths allowing each agent to reach its goal without collisions~\cite{Stern}. 
In this work, we shall define agent $a_i$'s \textit{path} as $\pi_i = \langle \pi_{i,t}, \pi_{i,t+1}, \ldots, \pi_{i,t+n} \rangle$, namely a finite sequence of vertices $\pi_{i,h} \in V$ starting at some time $t$ and ending at $t+n$. Following $\pi_i$, the agent must either move to an adjacent vertex ($(\pi_{i,t}, \pi_{i,t+1}) \in E$) or not move ($\pi_{i,t+1} = \pi_{i,t}$).

MAPD extends the above one--shot setting to a time--extended setting by introducing tasks $\tau_j \in \mathcal{T}$, each specifying a pickup and a delivery vertex denoted as $s_j$ and $g_j$, respectively. A task has to be assigned to an agent that must execute it following a collision--free path from its initial location to $s_j$ and then from $s_j$ to $g_j$. A peculiar characteristic of this problem is that the set $\mathcal{T}$ is filled at runtime: a task can be added to the system at any (finite) time and from the moment it is added it becomes assignable to any agent. An agent is \textit{free} when it is currently not executing any task and \textit{occupied} when it is assigned to a task.
If an agent is free, it can be assigned to any task $\tau_j \in \mathcal{T}$, with the constraint that a task can be assigned to only one agent. When this happens, the task is removed from $\mathcal{T}$ and, when the agent completes its task eventually arriving at $g_j$, it returns free. A \textit{plan} is a set of paths, which are required to be \textit{collision--free}, namely any two agents cannot be in the same vertex or traverse the same edge at the same time. Each action (movement to an adjacent vertex or wait) lasts one time step. Solving MAPD means finding a minimum--cost plan to complete all the tasks in $\mathcal{T}$. Cost usually takes one of two possible definitions. The \textit{service time} is the average number of time steps needed to complete each task $\tau_j$, measured as the time elapsed from $\tau_j$'s arrival to the time an agent reaches $g_j$. The \textit{makespan}, instead, is the earliest time step at which all the tasks are completed. Being MAPD a generalization of MAPF, it is NP--hard to solve optimally with any of the previous cost functions~\cite{Yu&LaValle,Surynek}.

Recent research focused on how to compute solutions of the above problems which are robust to delays, namely to runtime events blocking agents at their current vertices for one or more time steps, thus slowing down the paths execution. The MAPF literature provides two notions of robustness, which we will exploit in this paper. The first one is that of $k$--robustness~\cite{Atzmon_k_JAIR,chen2021symmetry}. A plan is $k$--robust iff it is collision--free and remains so when at most $k$ delays for each agent occur. To create $k$--robust plans, an algorithm should ensure that, when an agent leaves a vertex, that vertex is not occupied by another agent for at least $k$ time steps. In this way, even if the first agent delays $k$ times, no collision can occur. The second one is called $p$--robustness~\cite{Atzmon}. Assume that a fixed probability $p_d$ of any agent being delayed at any time step is given and that delays are independent of each other. Then, a plan is $p$--robust iff the probability that it will be executed without a collision is at least $p$. Differently from $k$--robustness, this notion provides a probabilistic guarantee.

Robustness for MAPD problems has been less studied. One notion proposed in~\cite{Koenig} and called \textit{long--term robustness} is actually a \textit{feasibility} property that guarantees that a finite number of tasks will be completed in a finite time. Authors show how a sufficient condition to have long--term robustness is to ensure that a MAPD instance is \emph{well--formed}. This amounts to require that (i) the number of tasks is finite; (ii) there are as much endpoints as agents, where endpoints are vertices designated as rest locations at which agents might not interfere with any other moving agent; (iii) for any two endpoints, there exists a path between them that traverses no other endpoints.

In this work, we leverage the above concepts to extend $k$-- and $p$--robustness to long--term MAPD settings. To do so, we will focus on a current state--of--the--art algorithm for MAPD, Token Passing (TP)~\cite{Koenig}. This algorithm follows an online and decentralized approach that, with respect to the centralized counterparts, trades off optimality to achieve an affordable computational cost in real--time long--term settings. We report it in Algorithm~\ref{alg:tp}. The \textit{token} is a shared block of memory containing the current agents' paths $\pi_{i}$s, the current task set $\mathcal{T}$, and the current assignment of tasks to the agents. The token is initialized with paths in which each agent $a_i$ rests at its initial location $loc(a_i)$ (line~\ref{TP:init}). At each time step, new tasks might be added to $\mathcal{T}$ (line~\ref{TP:newtasks}). When an agent has reached the end of its path in the token, it becomes free and requests the token (at most once per time step). The token is sent in turn to each requesting agent (line~\ref{TP:assigntoken}) and the agent with the token assigns itself (line~\ref{TP:assignagenttask}) to the task $\tau$ in $\mathcal{T}$ whose pickup vertex is closest to its current location (line~\ref{TP:mindist}), provided that no other path already planned (and stored in the token) ends at the pickup or delivery vertex of such task (line~\ref{TP:noends}). The distance between the current location $\textit{loc}(a_{i})$ of agent $a_{i}$ and the pickup location $s_{j}$ of a task is calculated using a (possibly approximated) function $h$ (for the grid environments of our experiments we use the Manhattan distance). The agent then computes a collision--free path from its current position to the pickup vertex, then from there to the delivery vertex, and finally it eventually rests at the delivery vertex (line~\ref{TP:pathplan}). Finally, the agent releases the token (line~\ref{TP:release}) and everybody moves one step on its path (line~\ref{TP:move}). If $a_i$ cannot find a feasible path it stays where it is (line~\ref{TP:update1}) or it calls the function \textit{Idle} to compute a path to an endpoint in order to ensure long--term robustness (line~\ref{TP:update2}).
\begin{algorithm}[t!]
\begin{small}
\caption{Token Passing}
\label{alg:tp}
\SetAlgoLined
 initialize \textit{token} with path $\langle \textit{loc}(a_i) \rangle $ for each agent $a_i$ ($\textit{loc}(a_i)$ is $a_{i}$'s current (eventually initial) location)\;\label{TP:init}
 \While{true}
 {
  add new tasks, if any, to the task set $\mathcal{T}$\;\label{TP:newtasks}
 \While{agent $a_i$ exists that requests token}
 {
  /* \textit{token} assigned to $a_i$ and $a_i$ executes now */\;\label{TP:assigntoken}
  $\mathcal{T'} \leftarrow \{\tau_j \in \mathcal{T} \mid$ no path in \textit{token} ends in $s_j$ or $g_j$\}\;\label{TP:noends}
  \uIf{$\mathcal{T'} \neq \{ \}$}
  {
   $\tau \leftarrow \argmin_{\tau_j \in \mathcal{T'}}$ $h(\textit{loc}(a_i),s_j)$\;\label{TP:mindist}
   assign $a_i$ to $\tau$\;\label{TP:assignagenttask}
   remove $\tau$ from $\mathcal{T}$\;
   update $a_i$’s path in \textit{token} with the path returned by $\textit{PathPlanner}(a_i, \tau, \textit{token})$\;\label{TP:pathplan}
  }
  \uElseIf{no task $\tau_j \in \mathcal{T}$ exists with $g_j = \textit{loc}(a_i)$}
  {
   update $a_i$’s path in \textit{token} with the path $\langle \textit{loc}(a_i) \rangle$\;\label{TP:update1}
  }
  \Else
  {
   update $a_i$’s path in \textit{token} with $\textit{Idle}(a_i, \textit{token})$;\label{TP:update2}
  }
  /* $a_i$ returns \textit{token} to system */\;\label{TP:release}
 }
 agents move on their paths in \textit{token} for one time step\;\label{TP:move}
 }
 \end{small}
\end{algorithm}

Note that other dynamic and online settings, different from ours, have been considered for MAPF and MAPD. For example,~\cite{svancara2019aaai} introduces a setting in which the set of agents is not fixed, but agents can enter and leave the system,~\cite{ma2021icaps} proposes an insightful comparison of online algorithms that can be applied to the aforementioned setting, and ~\cite{shahar2021safe} studies a related problem where the actions have uncertain costs.
\section{MAPD with Delays}\label{mapd_delays}

Delays are typical problems in real applications of MAPF and MAPD and may have multiple causes. 
For example, robots can slow down due to some errors occurring in the sensors used for localization and coordination~\cite{Khalastchi2019FaultDA}. Moreover, real robots are subject to physical constraints, like minimum turning radius, maximum velocity, and maximum acceleration, and, although algorithms exists to convert time--discrete MAPD plans into plans executable by real robots~\cite{Ma_kin}, small differences between models and actual agents may still cause delays. Another source of delays is represented by anomalies happening during path execution and caused, for example, by partial or temporary failures of some agent~\cite{anomaly}.

We define the problem of \textit{MAPD with delays} (\textit{MAPD--d}) as a MAPD problem (see Section~\ref{Related Works}) where the execution of the computed paths $\pi_i$ can be affected, at any time step $t$, by delays represented by a time--varying set $\mathcal{D}(t) \subseteq A$. Given a time step $t$, $\mathcal{D}(t)$ specifies the subset of agents that will delay the execution of their paths, lingering at their currently occupied vertices at time step $t$. An agent could be delayed for several consecutive time steps, but not for indefinitely long in order to preserve well--formedness (see next section). The temporal realization of $\mathcal{D}(t)$ is unknown when planning paths, so a {MAPD--d} instance is formulated as a MAPD one: no other information is available at planning time. The difference lies in how the solution is built: in MAPD--d we compute solutions accounting for robustness to delays that might happen at runtime.

More formally, delays affect each agent's execution trace.
Agent $a_i$'s \textit{execution trace} $e_i = \langle e_{i,0}, e_{i,1}, ..., e_{i,m} \rangle$\footnote{For simplicity and w.l.o.g., we consider a path and a corresponding execution trace starting from time step $0$.} for a given path $\pi_i = \langle \pi_{i,0}, \pi_{i,1}, \ldots , \pi_{i,n} \rangle$ corresponds to the actual sequence of $m$ ($m \geq n$) vertices traversed by $a_i$ while following $\pi_i$ and accounting for possible delays. Let us call $\textit{idx}(e_{i,t})$ the index of $e_{i,t}$ (the vertex occupied by $a_i$ at time step $t$) in $\pi_i$. Given that $e_{i,0} = \pi_{i,0}$, the execution trace is defined, for $t>0$, as:
$$e_{i,t}=
\begin{cases}
e_{i,t-1}& \text{if } a_i\in \mathcal{D}(t)\\
\pi_{i,h} \mid h = \textit{idx}(e_{i,t-1}) + 1 & \text{otherwise} 
\end{cases}$$
An execution trace terminates when $e_{i,m} = \pi_{i,n}$ for some $m$.

Notice that, if no delays are present (that is, $\mathcal{D}(t)=\{ \}$ for all $t$) then the execution trace $e_i$ exactly mirrors the path $\pi_i$ and, in case this is guaranteed in advance, the MAPD--d problem becomes \emph{de facto} a regular MAPD problem. In general, such a guarantee is not given and solving a MAPD-d problem opens the issue of computing collision--free tasks--fulfilling MAPD paths (optimizing service time or makespan) characterized by some level of robustness to delays.

The MAPD-d problem reduces to the MAPD problem as a special case, so the MAPD--d problem is NP-hard.

\subsection{Well-formedness of MAPD--d}
In principle, if a problem instance is well--formed, delays will not affect its feasibility (this property is also called long--term robustness, namely the guarantee that a finite number of tasks will be completed in a finite time, see Section~\ref{Related Works}). Indeed, well--formedness is given by specific topological properties of the environment and delays, by their definition, are not such a type of feature. There is, however, an exception to this argument corresponding to a case where a delay does cause a modification of the environment, eventually resulting in the loss of well--formedness and, in turn, of feasibility. This is the case where an agent is delayed indefinitely and cannot move anymore (namely when the agent is in $\mathcal{D}(t)$ for all $t \geq T$ for a given time step $T$). In such a situation, the agent becomes a new obstacle, potentially blocking a path critical for preserving the well--formedness. The assumption that an agent cannot be delayed indefinitely made in the previous section ensures the well-formedness of MAPD--d instances. More precisely, a MAPD--d instance is well--formed when, in addition to requirements (i)--(iii) from Section~\ref{Related Works}, it satisfies also: (iv)~any agent cannot be in $\mathcal{D}(t)$ forever (i.e., for all $t \geq T$ for a given $T$).

In a real context, condition (iv) amounts to removing or repairing the blocked agents. For instance, if an agent experiences a permanent fail, it will be removed (in this case its incomplete task will return in the task set and at least one agent must survive in the system) or repaired after a finite number of time steps. This guarantees that the well--formedness of a problem instance is preserved (or, more precisely, that it is restored after a finite time).

\subsection{A MAPD--d baseline: TP with replanning}
Algorithms able to solve well--formed MAPD problems, like TP, are in principle able to solve well--formed MAPD--d problems as well. The only issue is that these algorithms would return paths that do not consider possible delays occurring during execution. Delays cause paths to possibly collide, although they did not at planning time. (Note that, according to our assumptions, when an agent is delayed at time step $t$, there is no way to know for how long it will be delayed.)

In order to have a baseline to compare against the algorithms we propose in the next section, we introduce an adaptation of TP allowing it to work also in the presence of delays. Specifically, we add to TP a replanning mechanism that works as follows: when a collision is detected between agents following their paths, the token is assigned to one of the colliding agents to allow replanning of a new collision--free path. This is a modification of the original TP mechanism where the token can be assigned only to free agents that have reached the end of their paths (see Algorithm~\ref{alg:tp}). To do this, we require the token to include also the current execution traces of the agents.

\begin{algorithm}[t!]
\begin{small}
\caption{TP with replanning}
\label{alg:tp-rec}
\SetAlgoLined
 initialize \textit{token} with the (trivial) path $\langle \textit{loc}(a_i) \rangle $ for each agent $a_i$\;
 \While{true}
 {
  add new tasks, if any, to the task set $\mathcal{T}$\;
  $\mathcal{R} \leftarrow \textit{CheckCollisions}(\textit{token})$\;\label{TPR:checkcol}
  \ForEach{agent $a_i$ in $\mathcal{R}$}
  {
   retrieve task $\tau$ assigned to $a_i$\;\label{TPR:retrievetask}
   $\pi_i \leftarrow \textit{PathPlanner}(a_i, \tau, \textit{token})$\;\label{TPR:pathplan}
   \uIf{$\pi_i$ is not null}
   {
    update $a_i$’s path in \textit{token} with $\pi_i$\;
   }
   \Else
   {
    recovery from deadlocks;\label{TPR:recdead}
   }
  }
 \While{agent $a_i$ exists that requests token}
 {
  proceed like in Algorithm \ref{alg:tp} (lines~\ref{TP:assigntoken}-\ref{TP:release})\;
 }
 agents move along their paths in \textit{token} for one time step (or stay at their current position if delayed)\;

 }
 \end{small}
\end{algorithm}

\begin{figure}[!hbtp]
  \centering
  \includegraphics[width=0.8\linewidth]{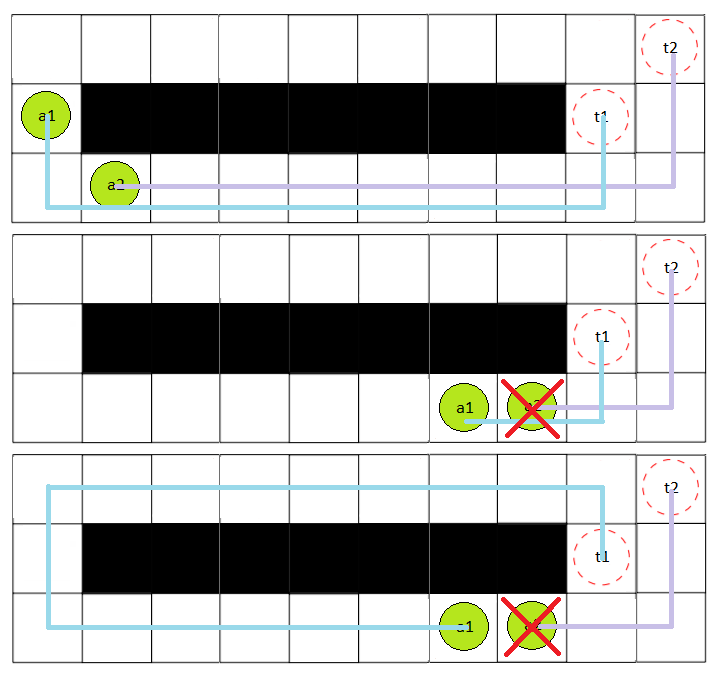}
  \caption{An example of TP with replanning. The figure shows a grid environment with two agents and two tasks at different time steps. Initially (top), the agents plan their paths without collisions. At time steps $6$ and $7$ (middle) $a_2$ is delayed and at time step $7$ a collision is detected in the token. Then, $a_1$ regains the token and replans (bottom).}
  \label{fig:delay_causes_replan}
\end{figure}

Algorithm~\ref{alg:tp-rec} reports the pseudo--code for this baseline method that we call TP with replanning. At the current time step a collision is checked using the function \textit{CheckCollisions} (line~\ref{TPR:checkcol}): a collision occurs at time step $t$ if an agent $a_{i}$ wants to move to the same vertex to which another agent $a_{j}$ wants to move or if $a_{i}$ and $a_{j}$ want to swap their locations on adjacent vertices. For example, this happens when $a_{j}$ is delayed at $t$ or when one of the two agents has been delayed at an earlier time step. The function returns the set $\mathcal{R}$ of non--delayed colliding agents that will try to plan new collision--free paths (line~\ref{TPR:pathplan}). The \textit{PathPlanner} function considers a set of constraints to avoid conflicts with the current paths of other agents in the token. A problem may happen when multiple delays occur at the same time; in particular situations, two or more agents may prevent each other to follow the only paths available to complete their tasks. In this case, the algorithm recognizes the situation and implements a deadlock recovery behavior. In particular, although with our assumptions agents cannot be delayed forever, we plan short collision--free random walks for the involved agents in order to speedup the deadlock resolution (line~\ref{TPR:recdead}). An example of execution of TP with replanning is depicted in Figure~\ref{fig:delay_causes_replan}.

\section{Algorithms for MAPD with delays}\label{S:algorithms}
  
In this section we present two algorithms, $k$--TP and $p$--TP, able to plan paths that solve MAPD-d problem instances with some guaranteed degree of robustness in face of delays. In particular, $k$--TP provides a deterministic degree of robustness, while $p$--TP provides a probabilistic degree of robustness. For developing these two algorithms, we took inspiration from the corresponding concepts of $k$-- and $p$--robustness for MAPF that we outlined in Section~\ref{Related Works}.
  
\subsection{$k$--TP Algorithm}

A $k$--\textit{robust} solution for MAPD--d is a plan which is guaranteed to avoid collisions due to at most $k$ consecutive delays for each agent, not only considering the paths already planned but also those planned in the future. (By the way, this is one of the main differences between our approach and the robustness for MAPF.) As we have discussed in Section~\ref{mapd_delays}, TP with replanning (Algorithm~\ref{alg:tp-rec}) can just react to the occurrence of delays once they have been detected. The $k$--TP algorithm we propose, instead, plans in advance considering that delays may occur, in the attempt of avoiding replanning at runtime. The algorithm is defined as an extension of TP with replanning, so it is able to solve all well--formed MAPD--d problem instances. A core difference is an additional set of constraints enforced during path planning.

The formal steps are reported in Algorithm~\ref{alg:k-tp}. A new path $\pi_i$, before being added to the token, is used to generate the constraints (the $k$--extension of the path, also added to the token, lines~\ref{kTP:extension1} and~\ref{kTP:extension2}) representing that, at any time step $t$, any vertex in $$\{ \pi_{i,t - k}, \ldots, \pi_{i,t-1}, \pi_{i,t}, \pi_{i,t+1}, \ldots,\pi_{i,t + k} \}$$ should be considered as an obstacle (at time step $t$) by agents planning later. In this way, even if agent $a_i$ or agent $a_j$ planning later are delayed up to $k$ times, no collision will occur. For example, if $\pi_{i} = \langle v_{1}, v_{2}, v_{3} \rangle$, the $1$-extension constraints will forbid any other agent to be in $\{ v_{1}, v_{2} \}$ at the first time step, in $\{ v_{1}, v_{2}, v_{3} \}$ at the second time step, in $\{ v_{2}, v_{3} \}$ at the third time step, and in $\{ v_{3} \}$ at the fourth time step.

\begin{algorithm}[!tbp]
\begin{small}
\caption{$k$-TP}
\label{alg:k-tp}
\SetAlgoLined
 initialize \textit{token} with the (trivial) path $\langle loc(a_i) \rangle$ for each agent $a_i$\;
 \While{true}
 {
  add new tasks, if any, to the task set $\mathcal{T}$\;
  $\mathcal{R} \leftarrow \textit{CheckCollisions}(\textit{token})$\;
  \ForEach{agent $a_i$ in $\mathcal{R}$}
  {
   proceed like in Algorithm~\ref{alg:tp-rec}\ (lines~\ref{TPR:retrievetask}-\ref{TPR:recdead});
  }
  \While{agent $a_i$ exists that requests token}
  {
   /* token is assigned to $a_i$ and $a_i$ executes now */\;
   $\mathcal{T'} \leftarrow \{\tau_j \in \mathcal{T} \mid$ no path in \textit{token} ends in $s_j$ or in $g_j$\}\;
   \uIf{$\mathcal{T'} \neq \{ \}$}
   {
    $\tau \leftarrow \argmin_{\tau_j \in \mathcal{T'}}$ $h(\textit{loc}(a_{i}),s_{j})$\;
    assign $a_i$ to $\tau$\;
    remove $\tau$ from $\mathcal{T}$\;
    $\pi_i \leftarrow \textit{PathPlanner}(a_i, \tau, \textit{token})$\;
    \uIf{$\pi_i$ is not null}
    {
     update \textit{token} with $\textit{k-extension}(\pi_i, k)$\;\label{kTP:extension1}
    }
   }
   \uElseIf{no task $\tau_j \in \mathcal{T}$ exists with $g_j = \textit{loc}(a_i)$}
   {
    update $a_i$’s path in \textit{token} with the path $\langle \textit{loc}(a_i) \rangle$\;
   }
   \Else
   {
    $\pi_i \leftarrow \textit{Idle}(a_i, \textit{token})$\;
    \uIf{$\pi_i$ is not null}
    {
     update \textit{token} with $\textit{k-extension}(\pi_i, k)$\;\label{kTP:extension2}
    }
   }
   /* $a_i$ returns \textit{token} to system */\;
 }
 agents move along their paths in \textit{token} for one time step (or stay at their current position if delayed)\;
 }
\end{small}
\end{algorithm}

The path of an agent added to the token ends at the delivery vertex of the assigned task, so the space requested in the token to store the path and the corresponding $k$--extension constraints is finite, for finite $k$. Note that, especially for large values of $k$, it may happen that a sufficiently robust path for an agent $a_i$ cannot be found at some time step; in this case, $a_i$ simply returns the token and tries to replan at the next time step. The idea is that, as other agents advance along their paths, the setting becomes less constrained and a path can be found more easily. Clearly, since delays that affect the execution are not known beforehand, replanning is still necessary in those cases where an agent gets delayed for more than $k$ consecutive time steps. 

\subsection{$p$--TP Algorithm}

The idea of $k$--robustness considers a fixed value $k$ for the guarantee, which could be hard to set: if $k$ is too low, plans may not be robust enough and the number of (possibly costly) replans could be high, while if $k$ is too high, it will increase the total cost of the solution with no extra benefit (see Section~\ref{S:experiments} for numerical data supporting these claims).

An alternative approach is to resort to the concept of $p$--robustness. A $p$--\textit{robust} plan guarantees to keep collision probability below a certain threshold $p$ ($0 \leq p \leq 1$). In a MAPD setting, where tasks are not known in advance, a plan could quickly reach the threshold with just few paths planned, so that no other path can be added to it until the current paths have been executed. Our solution to avoid this problem is to impose that only the collision probability of \emph{individual} paths should remain below the threshold $p$, not of the whole plan. s discussed in~\cite{wagner2017path}, this might also be a method to ensure a notion of fairness among agents.

We thus need a way to calculate the collision probability for a given path. We adopt a model based on Markov chains~\cite{levin2017markov}. Assuming that the probability that any agent is delayed at any time step is fixed and equal to $p_{d}$, we model agent $a_i$'s execution trace $ e_{i}$ (corresponding to a path $\pi_{i}$) with a Markov chain, where the transition matrix $P$ is such that with probability $p_d$ the agent remains at the current vertex and with probability $1 - p_d$ advances along $\pi_{i}$. We also assume that transitions along chains of different agents are independent. (This simplification avoids that delays for one agent propagate to other agents, which could be problematic for the model~\cite{wagner2017path}, while still providing an useful proxy for robustness.)

This model is leveraged by our $p$--TP algorithm reported as Algorithm~\ref{alg:p-tp}. The approach is again an extension of TP with replanning, so also in this case we are able to solve any well--formed MAPD instance. Here, one difference with the basic algorithms is that before inserting a new path $\pi_{i}$ in the token, the Markov chain model is used to calculate the collision probability $\textit{cprob}_{\pi_i}$ between $\pi_{i}$ and the paths already in the token (lines~\ref{pTP:cprob1} and~\ref{pTP:cprob2}). Specifically, the probability distribution for the vertex occupied by an agent $a_{i}$ at the beginning of a path $\pi_{i} = \langle \pi_{i,t}, \pi_{i,t+1}, \ldots , \pi_{i,t+n} \rangle$ is given by a (row) vector $s_0$ with length $n$ that has every element set to $0$ except that corresponding to the vertex $\pi_{i,t}$, which is $1$. The probability distribution for the location of an agent at time step $t+j$ is given by $s_0P^j$ (where $P$ is the transition matrix defined above). For example, in a situation with $3$ agents and $4$ vertices ($v_{1}, v_{2}, v_{3}, v_{4}$), the probability distributions at a given time step $t$ for the locations of agents $a_{1}$, $a_{2}$, and $a_{3}$ could be $\langle 0.6, 0.2, 0.1, 0.1\rangle$, $\langle 0.3, 0.2, 0.2, 0.3\rangle$, and $\langle 0.5, 0.1, 0.3, 0.1\rangle$, respectively. Then, for any vertex traversed by the path $\pi_{i}$, we calculate its collision probability as $1$ minus the probability that all the other agents are not at that vertex at that time step multiplied by the probability that the agent is actually at that vertex at the given time step. Following the above example, the collision probability in $v_{1}$ for agent $a_{1}$ at $t$ (i.e., the  probability that at least one of the other agents is at $v_{1}$ at $t$) is calculated as $ [1-(1-0.3)\cdot(1-0.5) ]\cdot0.6 = 0.39$.
The collision probabilities of all the vertices along the path are summed to obtain the collision probability $\textit{cprob}_{\pi_i}$ for the path $\pi_{i}$. If this probability is above the threshold $p$ (lines~\ref{pTP:thresholdp1} and~\ref{pTP:thresholdp2}), the path is rejected and a new one is calculated. If an enough robust path is not found after a fixed number of rejections \textit{itermax}, the token is returned to the system and the agent will try to replan at the next time step (as other agents advance along their paths, chances of collisions could decrease). 

\begin{algorithm}[!tbp]
\begin{small}
\caption{$p$-TP}
\label{alg:p-tp}
\SetKw{Kwbreak}{break}
\SetAlgoLined
 initialize \textit{token} with path $\langle loc(a_i) \rangle$ for each agent $a_i$\;
 \While{true}
 {
  add new tasks, if any, to the task set $\mathcal{T}$\;
  $\mathcal{R} \leftarrow \textit{CheckCollisions}(\textit{token})$\;
  \ForEach{agent $a_i$ in $\mathcal{R}$}
  {
   proceed like in Algorithm \ref{alg:tp-rec} (lines 7 - 13)\;
  }
  \While{agent $a_i$ exists that requests token}
  {
   /* token assigned to $a_i$ and $a_i$ executes now */\;
   $\mathcal{T'} \leftarrow \{\tau_j \in \mathcal{T} \mid$ no path in \textit{token} ends in $s_j$ or in $g_j$\}\;
   \uIf{$\mathcal{T'} \neq \{ \}$}
   {
    $\tau \leftarrow \argmin_{\tau_j \in \mathcal{T'}}$ $h(\textit{loc}(a_i),s_j)$\;
    assign $a_i$ to $\tau$\;
    remove $\tau$ from $\mathcal{T}$\;
    $j \leftarrow 0$\;
    \While{$j < \textit{itermax}$}
    {
     $\pi_i \leftarrow \textit{PathPlanner}(a_i, \tau, \textit{token})$\;
     $\textit{cprob}_{\pi_i} \leftarrow \textit{MarkovChain}(\pi_i,\textit{token})$\;\label{pTP:cprob1}
     \uIf{$\textit{cprob}_{\pi_i} < p$} 
     {\label{pTP:thresholdp1} 
      update $a_i$’s path in \textit{token} with $\pi_i$\;
      \Kwbreak{}
     }
     $j \leftarrow j+1$;
    }
   }
   \uElseIf{no task $\tau_j \in \mathcal{T}$ exists with $g_j = \textit{loc}(a_i)$}
   {
    update $a_i$’s path in \textit{token} with the path $\langle \textit{loc}(a_i) \rangle$\;
   }
   \Else
   {
    $j \leftarrow 0$\;
    \While{$j < \textit{itermax}$}
    {
     $\pi_i \leftarrow \textit{Idle}(a_i, \textit{token})$\;
     $\textit{cprob}_{\pi_i} \leftarrow \textit{MarkovChain}(\pi_i,\textit{token})$\;\label{pTP:cprob2}
     \uIf{$\textit{cprob}_{\pi_i} < p$}
     {\label{pTP:thresholdp2}
      update $a_i$’s path in \textit{token} with $\pi_i$\;
      \Kwbreak{}
     }
     $j \leftarrow j+1$\;
    }
   }
   /* $a_i$ returns \textit{token} and system executes now */\;
 }
  agents move along their paths in token for one time step (or stay at their current position if delayed)\;
 }
 \end{small}
\end{algorithm}

Also for $p$--TP, since the delays are not known beforehand, replanning is still necessary. Moreover, we need to set the value of $p_d$, with which we build the probabilistic guarantee according to the specific application setting. We deal with this in the next section.
\section{Experimental Results}\label{S:experiments}
  
\subsection{Setting}
\begin{figure}[t!]
  \centering
  \includegraphics[width=0.8\linewidth]{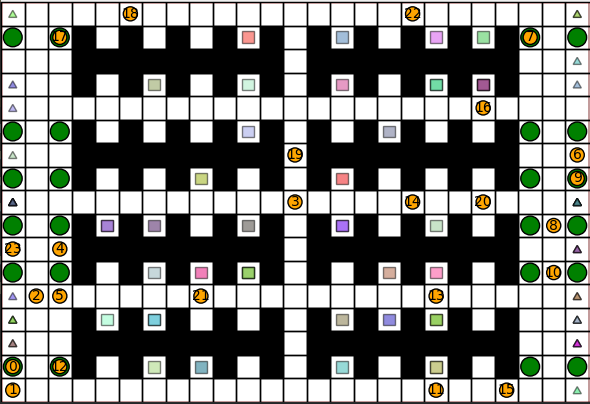}
  \caption{Large warehouse with 24 agents, obstacles (black), pickup (colored squares) and delivery (triangles) vertices, and endpoints (green circles)}
  \label{fig:large_warehouse}
\end{figure}

Our experiments are conducted on a 3.2 GHz Intel Core i7 8700H laptop with 16 GB of RAM. We tested our algorithms in two warehouse 4--connected grid environments where the effects of delays can be significant: a small one, $15 \times 13$ units, with $4$ and $8$ agents, and a large one, $25 \times 17$, with $12$ and $24$ agents (Figure~\ref{fig:large_warehouse}). (Environments of similar size have been used in~\cite{Koenig}.)
At the beginning, the agents are located at the endpoints. We create a sequence of $50$ tasks choosing the pickup and delivery vertices uniformly at random among a set of predefined vertices. The arrival time of each task is determined according to a Poisson distribution~\cite{poisson}. We test $3$ different arrival frequencies $\lambda$ for the tasks: $0.5$, $1$, and $3$ (since, as discussed later, the impact of $\lambda$ on robustness is not relevant, we do not show results for all values of $\lambda$). During each run, $10$ delays per agent are randomly inserted and the simulation ends when all the tasks have been completed.

We evaluate $k$--TP and $p$--TP against the baseline TP with replanning (to the best of our knowledge, we are not aware of any other algorithm for finding robust solutions to MAPD--d). For $p$--TP we use two different values for the parameter $p_d$, $0.02$ and $0.1$, modeling a low and a higher probability of delay, respectively. (Note that this is the expected delay probability used to calculate the robustness of a path and could not match with the delays actually observed.) For planning paths of individual agents (\textit{PathPlanner} in the algorithms), we use an A* path planner with Manhattan distance as heuristic. 

Solutions are evaluated according to the makespan (i.e., the earliest time step at which all tasks are completed, see Section~\ref{Related Works}). (Results for the service time are qualitatively similar and are not reported here.) We also consider the number of replans performed during execution and the total time required by each simulation (including time for both planning and execution). The reported results are averages over $100$ randomly restarted runs. All algorithms are implemented in Python and the code is publicly available at an online repository\footnote{Link hidden to keep anonymity.}.


\subsection{Results}
\label{S:results}
Results relative to small warehouse are shown in Tables~\ref{tab:small0.5} and~\ref{tab:small3} and those relative to large warehouse are shown in Tables~\ref{tab:mid0.5} and~\ref{tab:mid3}. For the sake of readability, we do not report the standard deviation in tables. Standard deviation values do not present any evident oddity and support the conclusions about the trends reported below. 

\begin{table}[t!]
  \caption{Small warehouse, $\lambda= 0.5$, and $10$ delays per agent}
  \label{tab:small0.5}
  \bgroup \def\arraystretch{1.5}
  \resizebox{\columnwidth}{!}{\begin{tabular}{|c|c|c|c|c|c|c|c|}
    \hline
    \multicolumn{2}{|c|}{} &\multicolumn{3}{|c|}{$\ell = 4$} &\multicolumn{3}{|c|}{$\ell = 8$}\\
     \cline{3-8}
     \multicolumn{2}{|c|}{$k$ or $p$}  & makespan & \# replans & runtime [s] & makespan & \#replans & runtime [s]\\\hline
  \multirow{5}{*}{\rotatebox{90}{\scriptsize{$k$-TP}}}  & 0 & \textbf{364.88} & 7.26 & \textbf{0.85} & \textbf{234.59} & 16.04 & \textbf{2.11} \\\cline{2-8}
& 1 & 374.48 & 1.4 & 0.91 & 240.69 & 3.85 & 2.27 \\\cline{2-8}
& 2 & 390.82 & 0.1 & 1.16 & 241.14 & 0.73 & 2.15 \\\cline{2-8}
& 3 & 411.09 & 0.01 & 1.59 & 259.38 & 0.09 & 3.12 \\\cline{2-8}
& 4 & 436.12 & \textbf{0.0} & 2.0 & 278.33 & \textbf{0.04} & 4.49 \\\cline{2-8}
   \hline
   \multirow{5}{*}{\rotatebox{90}{\scriptsize{$p$-TP, $p_d= .1$}}}
   & 1 & \textbf{364.88} & 7.26 & 1.14 & \textbf{234.59} & 16.04 & 2.63 \\\cline{2-8}
& 0.5 & 369.5 & 6.29 & 1.81 & 237.27 & 12.59 & 5.0 \\\cline{2-8}
& 0.25 & 395.07 & 4.29 & 2.88 & 255.21 & 5.63 & 6.11 \\\cline{2-8}
& 0.1 & 409.17 & 2.9 & 3.16 & 268.99 & 3.23 & 6.32 \\\cline{2-8}
& 0.05 & 428.64 & 2.93 & 3.42 & 279.26 & 2.76 & 6.48 \\\cline{2-8}
   \hline
   \multirow{5}{*}{\rotatebox{90}{\scriptsize{$\quad p$-TP, $p_d = .02$}}}  
& 0.5 & 366.72 & 7.34 & 1.29 & 238.83 & 12.81 & 3.87 \\\cline{2-8}
& 0.25 & 378.42 & 6.8 & 1.57 & 236.21 & 10.21 & 4.38 \\\cline{2-8}
& 0.1 & 391.63 & 4.53 & 2.37 & 250.39 & 6.73 & 5.57 \\\cline{2-8}
& 0.05 & 405.53 & 3.51 & 2.66 & 256.24 & 4.25 & 5.34 \\\cline{2-8}
   \hline  
  \end{tabular}}
  \egroup
\end{table}

\begin{table}[t!]
  \caption{Small warehouse, $\lambda = 3$, and $10$ delays per agent}
  \label{tab:small3}
  \bgroup \def\arraystretch{1.5}
  \resizebox{\columnwidth}{!}{\begin{tabular}{|c|c|c|c|c|c|c|c|}
    \hline
    \multicolumn{2}{|c|}{} &\multicolumn{3}{|c|}{$\ell = 4$} &\multicolumn{3}{|c|}{$\ell = 8$}\\
     \cline{3-8}
     \multicolumn{2}{|c|}{$k$ or $p$}  & makespan & \# replans & runtime [s] & makespan & \# replans & runtime [s]\\\hline
  \multirow{5}{*}{\rotatebox{90}{\scriptsize{$k$-TP}}}  & 0 & \textbf{354.77} & 8.3 & \textbf{0.6} & \textbf{217.79} & 14.67 & 1.93 \\\cline{2-8}
& 1 & 363.22 & 1.47 & 0.77 & 219.87 & 4.01 & \textbf{1.81} \\\cline{2-8}
& 2 & 383.59 & 0.2 & 0.95 & 226.75 & 0.58 & 1.89 \\\cline{2-8}
& 3 & 400.77 & 0.01 & 1.33 & 250.23 & 0.12 & 3.02 \\\cline{2-8}
& 4 & 429.12 & \textbf{0.0} & 1.68 & 263.47 & \textbf{0.01} & 4.32 \\\cline{2-8}
   \hline
   \multirow{5}{*}{\rotatebox{90}{\scriptsize{$p$-TP, $p_d= .1$}}} & 1 & \textbf{354.77} & 8.3 & 0.86 & \textbf{217.79} & 14.67 & 2.53 \\\cline{2-8}
& 0.5 & 360.29 & 6.7 & 1.45 & 224.31 & 11.06 & 4.93 \\\cline{2-8}
& 0.25 & 381.98 & 5.12 & 2.3 & 245.24 & 6.46 & 5.83 \\\cline{2-8}
& 0.1 & 404.92 & 2.93 & 2.81 & 251.42 & 3.55 & 5.66 \\\cline{2-8}
& 0.05 & 417.04 & 2.65 & 3.05 & 262.73 & 3.65 & 6.11 \\\cline{2-8}
   \hline
   \multirow{5}{*}{\rotatebox{90}{\scriptsize{$\quad p$-TP, $p_d = .02$}}}  
& 0.5 & 358.14 & 8.05 & 1.25 & 219.58 & 13.19 & 3.61 \\\cline{2-8}
& 0.25 & 372.92 & 7.02 & 1.57 & 228.25 & 10.93 & 3.77 \\\cline{2-8}
& 0.1 & 380.31 & 4.41 & 2.12 & 233.97 & 6.89 & 4.65 \\\cline{2-8}
& 0.05 & 393.55 & 3.45 & 2.5 & 244.62 & 4.81 & 4.98 \\\cline{2-8}
   \hline  
  \end{tabular}}
  \egroup
\end{table}

The baseline algorithm, TP with replanning, appears twice in each table: as $k$--TP with $k=0$ (that is the basic implementation as in Algorithm~\ref{alg:tp-rec}) and as $p$--TP with $p_{d}=0.1$ and $p = 1$ (which accepts all paths). The two versions of the baseline return the same results in terms of makespan and number of replans (we use the same random seed initialization for runs with different algorithms), but the total runtime is larger in the case of $p$--TP, due to the overhead of calculating the Markov chains and the collision probability for each path. 

Looking at robustness, which is the goal of our algorithms, we can see that, in all settings, both $k$--TP and $p$--TP significantly reduce the number of replans with respect to the baseline. For $k$--TP, increasing $k$ leads to increasingly more robust solutions with less replans, and the same happens for $p$--TP when the threshold probability $p$ is reduced. However, increasing $k$ shows a more evident effect on the number of replans than reducing $p$. More robust solutions, as expected, tend to have a larger makespan, but the first levels of robustness ($k=1$, $p=0.5$) manage to reduce significantly the number of replans with a small or no increase in makespan. For instance, in Table~\ref{tab:mid3}, $k$--TP with $k=1$ decreases the number of replans of more than $75\%$ with an increase in makespan of less than $2\%$, with respect to the baseline.  Pushing towards higher degrees of robustness (i.e., increasing $k$ or decreasing $p$) tends to increase makespan significantly with diminishing returns in terms of number of replans, especially for $k$--TP.

\begin{table}[t!]
  \caption{Large warehouse, $\lambda = 0.5$, and $10$ delays per agent}
  \label{tab:mid0.5}
  \bgroup \def\arraystretch{1.5}
  \resizebox{\columnwidth}{!}{\begin{tabular}{|c|c|c|c|c|c|c|c|}
    \hline
    \multicolumn{2}{|c|}{} &\multicolumn{3}{|c|}{$\ell = 12$} &\multicolumn{3}{|c|}{$\ell = 24$}\\
     \cline{3-8}
     \multicolumn{2}{|c|}{$k$ or $p$}  & makespan & \# replans & runtime [s] & makespan & \# replans & runtime [s]\\\hline
  \multirow{5}{*}{\rotatebox{90}{\scriptsize{$k$-TP}}}  & 0 & 283.62 & 17.18 & \textbf{2.8} & 269.25 & 20.71 & 8.32 \\\cline{2-8}
& 1 & \textbf{276.7} & 3.88 & 3.27 & \textbf{264.96} & 5.37 & \textbf{5.78} \\\cline{2-8}
& 2 & 285.32 & 1.18 & 4.89 & 275.48 & 1.62 & 9.54 \\\cline{2-8}
& 3 & 304.05 & 0.24 & 7.54 & 300.55 & 0.4 & 15.55 \\\cline{2-8}
& 4 & 310.59 & \textbf{0.01} & 10.9 & 300.45 & \textbf{0.1} & 22.11 \\\cline{2-8}
   \hline
   \multirow{5}{*}{\rotatebox{90}{\scriptsize{$p$-TP, $p_d= .1$}}} & 1 & 283.62 & 17.18 & 4.12 & 269.25 & 20.71 & 11.2 \\\cline{2-8}
& 0.5 & 286.95 & 10.02 & 11.3 & 291.78 & 17.09 & 38.61 \\\cline{2-8}
& 0.25 & 305.13 & 5.38 & 17.26 & 313.63 & 9.59 & 58.95 \\\cline{2-8}
& 0.1 & 330.58 & 4.51 & 19.6 & 322.26 & 4.51 & 54.92 \\\cline{2-8}
& 0.05 & 337.33 & 3.56 & 20.27 & 348.89 & 3.89 & 57.24 \\\cline{2-8}
   \hline
   \multirow{5}{*}{\rotatebox{90}{\scriptsize{$\quad p$-TP, $p_d = .02$}}}  
& 0.5 & 289.86 & 14.51 & 7.41 & 290.05 & 20.3 & 28.74 \\\cline{2-8}
& 0.25 & 287.72 & 9.92 & 10.19 & 286.77 & 14.15 & 39.47 \\\cline{2-8}
& 0.1 & 311 & 6.53 & 13.76 & 304.24 & 8.94 & 49.04 \\\cline{2-8}
& 0.05 & 313.38 & 6.41 & 14.91 & 308.1 & 7.02 & 49.96 \\\cline{2-8}
   \hline  
  \end{tabular}}
  \egroup
\end{table}

\begin{table}[t!]
  \caption{Large warehouse, $\lambda = 3$, and $10$ delays per agent}
  \label{tab:mid3}
  \bgroup \def\arraystretch{1.5}
  \resizebox{\columnwidth}{!}{\begin{tabular}{|c|c|c|c|c|c|c|c|}
    \hline
    \multicolumn{2}{|c|}{} &\multicolumn{3}{|c|}{$\ell = 12$} &\multicolumn{3}{|c|}{$\ell = 24$}\\
     \cline{3-8}
     \multicolumn{2}{|c|}{$k$ or $p$}  & makespan & \# replans & runtime [s] & makespan & \# replans & runtime [s]\\\hline
  \multirow{5}{*}{\rotatebox{90}{\scriptsize{$k$-TP}}}  & 0 & 265.23 & 18.96 & \textbf{2.91} & 258.49 & 30.83 & \textbf{8.12} \\\cline{2-8}
& 1 & 269.78 & 4.22 & 3.28 & 254.56 & 8.98 & 9.81 \\\cline{2-8}
& 2 & 274.78 & 1.19 & 4.75 & 261.3 & 1.71 & 12.03 \\\cline{2-8}
& 3 & 279.02 & 0.18 & 7.31 & 273.56 & 0.59 & 19.43 \\\cline{2-8}
& 4 & 290.59 & \textbf{0.04} & 10.76 & 282.07 & \textbf{0.17} & 30.91 \\\cline{2-8}
   \hline
   \multirow{5}{*}{\rotatebox{90}{\scriptsize{$p$-TP, $p_d= .1$}}} & 1 & 265.23 & 18.96 & 4.16 & 258.49 & 30.83 & 10.78 \\\cline{2-8}
& 0.5 & 268.74 & 11.31 & 9.04 & 257.64 & 17.21 & 36.74 \\\cline{2-8}
& 0.25 & 298.01 & 7.39 & 14.58 & 287.75 & 9.96 & 48.14 \\\cline{2-8}
& 0.1 & 318.37 & 5.3 & 16.33 & 310.46 & 6.32 & 47.11 \\\cline{2-8}
& 0.05 & 331.1 & 3.83 & 16.83 & 334.06 & 4.42 & 47.62 \\\cline{2-8}
   \hline
   \multirow{5}{*}{\rotatebox{90}{\scriptsize{$\quad p$-TP, $p_d = .02$}}}  
& 0.5 & \textbf{259.64} & 12.47 & 7.22 & \textbf{247.76} & 20.47 & 26.21 \\\cline{2-8}
& 0.25 & 289.75 & 12.05 & 9.23 & 264.6 & 15.72 & 39.68 \\\cline{2-8}
& 0.1 & 280.07 & 6.78 & 11.59 & 290.65 & 9.88 & 42.76 \\\cline{2-8}
& 0.05 & 298.34 & 6.21 & 12.98 & 293.68 & 8.81 & 42.23 \\\cline{2-8}
   \hline  
  \end{tabular}}
  \egroup
\end{table}

Comparing $k$--TP and $p$--TP, it is clear that solutions produced by $k$--TP tend to be more robust at similar makespan (e.g., see $k$--TP with $k=1$ and $p$--TP with $p_{d}=.1$ and $p=0.5$ in Table~\ref{tab:small0.5}), and decreasing $p$ may sometimes lead to relevant increases in makespan. This suggests that our implementation of $p$--TP has margins for improvement: if the computed path exceeds the threshold $p$ we wait the next time step to replan, without storing any collision information extracted from the Markov chains; finding ways to exploit this information may lead to an enhanced version of $p$--TP (this investigation is left as future work). It is also interesting to notice the effect of $p_d$ in $p$--TP: a higher $p_d$ (which, in our experiments, amounts to overestimating the actual delay probability that, considering that runs last on average about $300$ time steps and there are $10$ delays per agent, is equal to $\frac{10}{300}=0.03$) leads to solutions requiring less replans, but with a noticeable increase in makespan. 

Considering runtimes, $k$--TP and $p$--TP are quite different. For $k$--TP, we see a trend similar to that observed for makespan: a low value of $k$ ($k=1$) often corresponds to a slight increase in runtime with respect to the baseline (sometimes even a decrease), while for larger values of $k$ the runtime may be much longer than the baseline. Instead, $p$--TP shows a big increase in runtime with respect to the baseline, that does not change too much with the values of $p$, at least for low values of $p$ ($p=0.1$, $p=0.05$). Finally, we can see how different task frequencies $\lambda$ have no significant impact on our algorithms, but higher frequencies have the global effect of reducing makespan tasks (which are always $50$ per run) are available earlier.

We repeat the previous experiments increasing the number of random delays inserted in execution to $50$ per agent, thus generating a scenario with multiple troubled agents. We show results for task frequency $\lambda = 1$ in Tables~\ref{tab:delays_small} and~\ref{tab:delays_mid}. Both algorithms significantly reduce the number of replans with respect to the baseline, reinforcing the importance of addressing possible delays during planning and not only during execution, especially when the delays can dramatically affect the operations of the agents, like in this case. The $k$--TP algorithm performs better than the $p$--TP algorithm, with trends similar to those discussed above. Note that, especially in the more constrained small warehouse (Table~\ref{tab:delays_small}), the big reduction in the number of replans produces a shorter runtime for $k$--TP with small values of $k$ wrt the baseline TP.

\begin{table}[t!]
  \caption{Small warehouse, $\lambda= 1$, and $50$ delays per agent}
  \label{tab:delays_small}
  \bgroup \def\arraystretch{1.5}
  \resizebox{\columnwidth}{!}{\begin{tabular}{|c|c|c|c|c|c|c|c|}
    \hline
    \multicolumn{2}{|c|}{} &\multicolumn{3}{|c|}{$\ell = 4$} &\multicolumn{3}{|c|}{$\ell = 8$}\\
     \cline{3-8}
     \multicolumn{2}{|c|}{$k$ or $p$}  & makespan & \# replans & runtime [s] & makespan & \# replans & runtime [s]\\\hline
  \multirow{5}{*}{\rotatebox{90}{\scriptsize{$k$-TP}}}  & 0 & 419.86 & 24.52 & 1.34 & 283.42 & 44.27 & 4.37 \\\cline{2-8}
& 1 & 424.1 & 8.77 & \textbf{0.87} & 283.69 & 18.35 & 3.21 \\\cline{2-8}
& 2 & 427.79 & 3.88 & 1.03 & 279.91 & 8.28 & \textbf{3.18} \\\cline{2-8}
& 3 & 445.52 & 1.27 & 1.46 & 303.73 & 4.7 & 3.66 \\\cline{2-8}
& 4 & 470.42 & \textbf{0.53} & 1.74 & 307.76 & \textbf{2.17} & 4.63 \\\cline{2-8}
   \hline
   \multirow{5}{*}{\rotatebox{90}{\scriptsize{$p$-TP, $p_d= .1$}}} & 1 & 419.86 & 24.52 & 1.71 & 283.42 & 44.27 & 5.64 \\\cline{2-8}
& 0.5 & 414.79 & 16.18 & 1.64 & 283.58 & 28.85 & 7.74 \\\cline{2-8}
& 0.25 & 430.99 & 11.83 & 2.46 & 294.97 & 15.42 & 8.03 \\\cline{2-8}
& 0.1 & 448.5 & 6.82 & 2.81 & 300.26 & 8.39 & 8.48 \\\cline{2-8}
& 0.05 & 458.92 & 5.68 & 2.91 & 309.03 & 5.77 & 7.38 \\\cline{2-8}
   \hline
   \multirow{5}{*}{\rotatebox{90}{\scriptsize{$\quad p$-TP, $p_d = .02$}}}  
& 0.5 & \textbf{407.29} & 18.47 & 1.46 & \textbf{271.96} & 32.15 & 5.44 \\\cline{2-8}
& 0.25 & 417.52 & 16.62 & 1.69 & 285.29 & 28.41 & 6.49 \\\cline{2-8}
& 0.1 & 430.55 & 12.5 & 2.26 & 290.72 & 17.75 & 7.14 \\\cline{2-8}
& 0.05 & 439.95 & 7.83 & 2.41 & 291.05 & 9.76 & 6.12 \\\cline{2-8}
   \hline  
  \end{tabular}}
  \egroup
\end{table}

\begin{table}[h]
  \caption{Large warehouse, $\lambda= 1$, and $50$ delays per agent}
  \label{tab:delays_mid}
  \bgroup \def\arraystretch{1.5}
  \resizebox{\columnwidth}{!}{\begin{tabular}{|c|c|c|c|c|c|c|c|}
    \hline
    \multicolumn{2}{|c|}{} &\multicolumn{3}{|c|}{$\ell = 12$} &\multicolumn{3}{|c|}{$\ell = 24$}\\
     \cline{3-8}
     \multicolumn{2}{|c|}{$k$ or $p$}  & makespan & \# replans & runtime [s] & makespan & \#replans & runtime [s]\\\hline
  \multirow{5}{*}{\rotatebox{90}{\scriptsize{$k$-TP}}}  & 0 & \textbf{310.51} & 42.8 & 4.83 & 317.23 & 66.53 & 12.66 \\\cline{2-8}
& 1 & 314.26 & 18.79 & \textbf{4.7} & \textbf{303.98} & 26.76 & \textbf{12.26} \\\cline{2-8}
& 2 & 321.13 & 9.43 & 5.98 & 316.6 & 18.29 & 16.56 \\\cline{2-8}
& 3 & 330.1 & 4.7 & 7.8 & 333.35 & 7.61 & 22.42 \\\cline{2-8}
& 4 & 345.93 & \textbf{2.98} & 11.26 & 336.2 & \textbf{4.7} & 28.49 \\\cline{2-8}
   \hline
   \multirow{5}{*}{\rotatebox{90}{\scriptsize{$p$-TP, $p_d= .1$}}} & 1 & \textbf{310.51} & 42.8 & 9.71 & 317.23 & 66.53 & 19.36 \\\cline{2-8}
& 0.5 & 330.24 & 28.99 & 19.26 & 319.39 & 38.59 & 48.64 \\\cline{2-8}
& 0.25 & 337.99 & 17.06 & 23.28 & 341.49 & 22.81 & 62.19 \\\cline{2-8}
& 0.1 & 355.03 & 10.16 & 25.25 & 368.1 & 13.19 & 63.77 \\\cline{2-8}
& 0.05 & 371.34 & 7.23 & 25.21 & 367.2 & 9.48 & 56.24 \\\cline{2-8}
   \hline
   \multirow{5}{*}{\rotatebox{90}{\scriptsize{$\quad p$-TP, $p_d = .02$}}}  
& 0.5 & 323.94 & 35.19 & 9.66 & 320.26 & 49.95 & 37.02 \\\cline{2-8}
& 0.25 & 326.09 & 27.45 & 11.6 & 339.79 & 40.87 & 56.83 \\\cline{2-8}
& 0.1 & 330.93 & 15.9 & 13.6 & 342.91 & 24.73 & 54.53 \\\cline{2-8}
& 0.05 & 350.39 & 15.67 & 15.11 & 345.79 & 20.17 & 55.37 \\\cline{2-8}
   \hline  
  \end{tabular}}
  \egroup
\end{table}

Finally, we run simulations in a even larger warehouse 4--con\-nected grid environment of size $25 \times 37$, with $50$ agents, $\lambda = 1$, $100$ tasks, and $10$ delays per agent. The same qualitative trends discussed above are observed also in this case. For example, $k$--TP with $k=2$ reduces the number of replans of $93\%$ with an increase of makespan of $5\%$ with respect to the baseline. The runtime of $p$--TP grows to hundreds of seconds, also with large values of $p$, suggesting that some improvements are needed. Full results are not reported here due to space constraints.

\section{Conclusion}

In this paper, we introduced a variation of the Multi-Agent Pickup and Delivery (MAPD) problem, called MAPD with delays (MAPD--d), which considers an important practical issue encountered in real applications: delays in execution. In a MAPD--d problem, agents must complete a set of incoming tasks (by moving to the pickup vertex of each task and then to the corresponding delivery vertex) even if they are affected by an unknown but finite number of delays during execution. We proposed two algorithms to solve MAPD--d, $k$--TP and $p$--TP, that are able to solve well--formed MAPD--d problem instances and provide deterministic and probabilistic robustness guarantees, respectively. Experimentally, we compared them against a baseline algorithm that reactively deals with delays during execution. Both $k$--TP and $p$--TP plan robust solutions, greatly reducing the number of replans needed with a small increase in solution makespan. $k$--TP showed the best results in terms of robustness--cost trade--off, but $p$--TP still offers great opportunities for future improvements.

Future work will address the enhancement of $p$--TP according to what we outlined in Section~\ref{S:results} and the experimental testing of our algorithms in real--world settings.

\bibliographystyle{ACM-Reference-Format} 
\bibliography{references}


\end{document}